\documentclass[10pt, conference, compsocconf]{IEEEtran}
\IEEEoverridecommandlockouts

\usepackage{cite}
\usepackage{amsmath,amssymb,amsfonts}
\usepackage{algorithmic}
\usepackage{graphicx}
\usepackage{textcomp}
\usepackage{booktabs}
\usepackage{makecell}
\usepackage{xcolor}
\def\BibTeX{{\rm B\kern-.05em{\sc i\kern-.025em b}\kern-.08em
    T\kern-.1667em\lower.7ex\hbox{E}\kern-.125emX}}
\begin{document}

\title{From Pixels to Reality: Physical-Digital Patch Attacks on Real-World Camera\\
{}
\thanks{This work was supported by the The Ministry of Economic Development of the Russian Federation in accordance with the subsidy agreement (agreement identifier 000000C313925P4H0002; grant No 139-15-2025-012).}
}

\author{\IEEEauthorblockN{Victoria Leonenkova}
\IEEEauthorblockA{\textit{Lomonosov MSU,} \\
Moscow, Russia \\
victoria.leonenkova@ \\
graphics.cs.msu.ru
}
\and
\IEEEauthorblockN{Ekaterina Shumitskaya}
\IEEEauthorblockA{\textit{AI Center, Lomonosov MSU,} \\
\textit{IAI MSU,}\\
Moscow, Russia \\
ekaterina.shumitskaya@ \\ 
graphics.cs.msu.ru}
\and
\IEEEauthorblockN{Dmitriy Vatolin}
\IEEEauthorblockA{\textit{AI Center, Lomonosov MSU,} \\
\textit{IAI MSU,}\\
Moscow, Russia \\
dmitriy@graphics.cs.msu.ru}
\and
\IEEEauthorblockN{Anastasia Antsiferova}
\IEEEauthorblockA{\textit{AI Center, Lomonosov MSU,} \\
\textit{IAI MSU,}\\
Moscow, Russia \\
aantsiferova@graphics.cs.msu.ru}
}

\maketitle

\begin{abstract}
This demonstration presents \textit{Digital-Physical Adversarial Attacks (DiPA)}, a new class of practical adversarial attacks against pervasive camera-based authentication systems, where an attacker displays an adversarial patch directly on a smartphone screen instead of relying on printed artifacts. This digital-only physical presentation enables rapid deployment, removes the need for total-variation regularization, and improves patch transferability in black-box conditions. DiPA leverages an ensemble of state-of-the-art face-recognition models (ArcFace, MagFace, CosFace) to enhance transfer across unseen commercial systems. Our interactive demo shows a \textit{real-time dodging attack} against a deployed face-recognition camera, preventing authorized users from being recognized while participants dynamically adjust patch patterns and observe immediate effects on the sensing pipeline. We further demonstrate DiPA’s superiority over existing physical attacks in terms of success rate, feature-space distortion, and reductions in detection confidence, highlighting critical vulnerabilities at the intersection of mobile devices, pervasive vision, and sensor-driven authentication infrastructures.
\end{abstract}


\begin{figure*}[htbp]
\centerline{\includegraphics[width=0.63\textwidth]{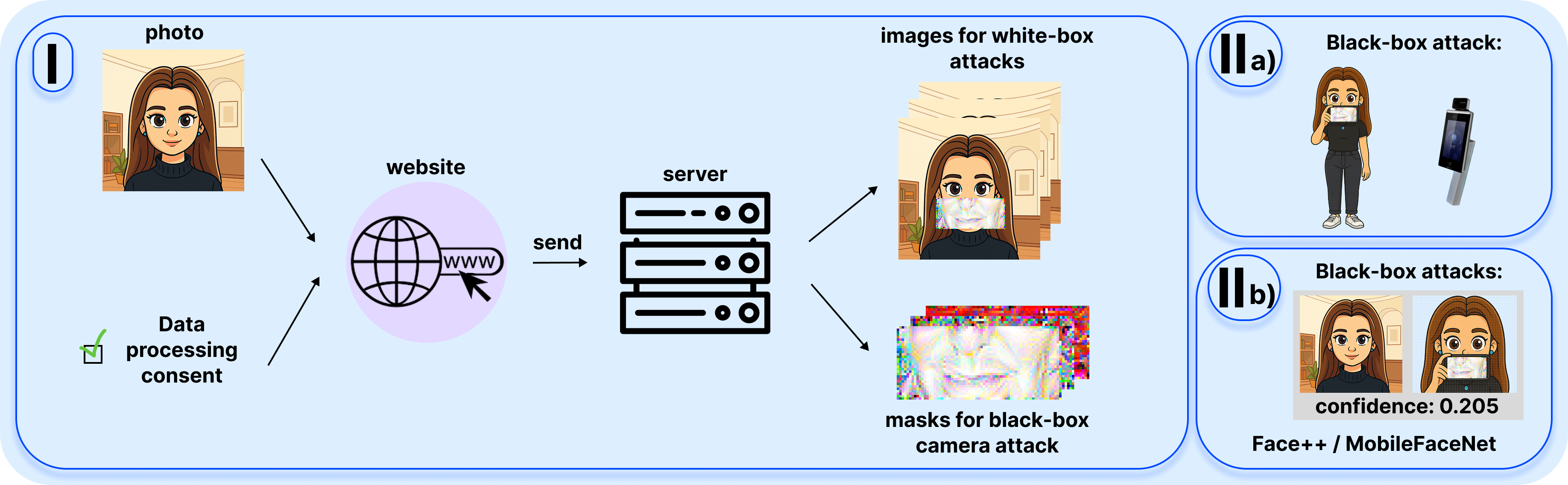}}
\caption{The overview of the proposed demonstration. The user uploads a photo to the server and receives a set of digital adversarial patches designed to cause dodging physical attack for a real camera in a completely black-box setting, along with images for white-box attacks on online face recognition models.}
\label{fig:first_figure}
\end{figure*}

\section{Introduction and Related Work}

Pervasive computing environments increasingly rely on vision-based authentication, particularly face recognition (FR), to enable frictionless access control for smart buildings, workplaces, homes, and public infrastructures. As FR modules become embedded in IoT devices, smart cameras, edge processors, and mobile systems, their security becomes a critical requirement for trustworthy pervasive computing ecosystems. Unfortunately, modern FR models—whether deployed on-device or in the cloud—remain vulnerable to adversarial attacks. Such attacks can be launched either digitally (e.g., manipulating input images before they reach the model) or physically (e.g., using adversarial objects or printed patterns visible to a camera).

Physical adversarial attacks are especially concerning in pervasive systems because they bypass digital safeguards and exploit the uncontrolled sensing pipeline, which includes illumination variations, sensor noise, optics, autofocus, and proprietary model preprocessing. Defense mechanisms for these attacks remain limited, and testing is often performed in purely digital setups that fail to represent complex real-world camera behavior.

Several digital~\cite{li2023sibling} and physical~\cite{zolfi2023adversarial,FaceOff} attacks on FR models have been proposed. However, their evaluations rarely include commercial, closed‑source camera systems—a setting highly relevant to deployed intelligent environments. Some early works incorporate cameras~\cite{8958122,pautov2019adversarial}, but real-world transferability is not their main target, and the adversarial patterns are typically printed, requiring careful color calibration and offering limited flexibility.

\textbf{In pervasive scenarios, users constantly carry a smartphone}, enabling both routine and potentially harmful interactions with camera systems. Surprisingly, no prior physical attack leverages the smartphone display itself as an adversarial surface, despite its ubiquity, mobility, brightness control, and dynamic rendering capabilities.

This demonstration presents \textbf{DiPA}, the first adversarial attack designed for digital-physical deployment on smartphone screens, enabling practical dodging attacks against real-world FR cameras. Our contributions include:

\begin{itemize}
\item \textbf{Novel attack modality for pervasive environments}: We introduce the first adversarial patch that is displayed directly on a smartphone screen and is immediately usable in real-world physical spaces, aligning with pervasive contexts where mobile devices interact continuously with sensing infrastructures.
\item \textbf{A method for generating high-resolution adversarial patches} that successfully transfer from surrogate FR models to commercial cameras, even in a strict black-box setting.
\item \textbf{A working end-to-end system}, accessible through a web interface, that allows participants to upload a picture and receive personalized adversarial patches for online and physical FR systems.
\item \textbf{Comprehensive evaluation} across commercial APIs and a real camera, demonstrating that DiPA and DiPA\textsubscript{TV} outperform prior physical and digital attacks in dodging performance.
\end{itemize}


\section{Approach}

Our work focuses on a \textit{dodging attack} designed to prevent the successful verification of an authorized user already enrolled in a face recognition (FR) system. Such attacks are highly relevant to pervasive environments where FR-enabled devices (e.g., smart doors, kiosks, embedded access-control terminals) operate autonomously at the edge and must handle uncontrolled user behavior and mobile-device interactions.

Let $\mathcal{F}: \mathbb{R}^{H \times W \times 3} \rightarrow \mathbb{R}^d$ denote an FR model that maps an input face image $x$ to an embedding vector. Given an enrolled identity with reference embedding $s = \mathcal{F}(x_{\text{ref}})$, the goal of the dodging attack is to generate an adversarial patch $p \in \mathbb{R}^{D \times D \times 3}$ that minimizes the cosine similarity between $\mathcal{F}(A(x,p))$---where $A(\cdot)$ is a patch-application operator---and the reference embedding $s$:
\begin{equation}
    p^* = \arg\min_{\mathbf{p}} 
    \sum_{i=1}^N \cos\!\left(
    \mathcal{F}_i\!\left(A\left(\mathbf{x}, \text{MedianPool}(\mathbf{p})\right)\right), 
    \mathcal{F}_i(\mathbf{x})
    \right),
\end{equation}
where $\{\mathcal{F}_i\}_{i=1}^N$ is the ensemble of surrogate FR models (ArcFace~\cite{deng2019arcface}, CosFace~\cite{wang2018cosface}, MagFace~\cite{meng2021magface}) and $\mathbf{x}$ is the attacker's face image.

Direct optimization against a real-world camera is infeasible due to the lack of gradients and strict limitations on the number of queries. Therefore, we rely on \textit{transferability} from surrogate models. Since these models accept low-resolution inputs ($3\times112\times112$), generating high-resolution physically deployable adversarial patches is challenging. To address this, we introduce a median-pooling mechanism that allows the patch to be trained at high resolution while still matching model constraints.

\subsection{Regularization Variants}

We study two versions of the proposed method:
\begin{itemize}
    \item \textbf{DiPA+TV}: Uses Total Variation (TV) regularization~\cite{mahendran2015understanding} to enforce spatial smoothness:
    \begin{equation}
        TV(p) = \sum_{i,j} \sqrt{(p_{i,j} - p_{i+1,j})^2 + (p_{i,j} - p_{i,j+1})^2}.
    \end{equation}
    \item \textbf{DiPA}: Removes TV regularization entirely. Because our patches are displayed digitally on smartphone screens (rather than printed), enforcing smoothness is less critical, and removing it increases the expressive power of the patch.
\end{itemize}


\subsection{Evaluation Across Digital--Physical and Real Pervasive Settings}

We compare DiPA with existing attacks on two online FR systems (Face++~\cite{faceplusplus}, MobileFaceNet~\cite{chen2018mobilefacenets}) and a commercial off-the-shelf FR camera operating as a strict black box.

For online models, we compute cosine similarity between embeddings of clean and patched images; lower values indicate stronger dodging performance.

For the real-world camera, we use two metrics:
\begin{enumerate}
    \item \textbf{Attack Success Rate (ASR)}:
    \begin{equation}
        ASR = \frac{1}{N}\sum_{i=1}^{N}
        \mathbb{I}\!\left(
        \mathcal{F}(A(x_i,p)) \neq \mathcal{F}(x_i)
        \right),
    \end{equation}
    where $N$ is the number of trials. The camera system reports the predicted identity $ \mathcal{F}(\cdot)$ directly, so this metric measures how often the presence of the smartphone screen with the adversarial patch causes the system to output an identity different from that of the authorized user.
    \item \textbf{Mean confidence of face presence}, which ensures the attack does not disrupt the face-detection module---an important property for pervasive deployments, where detection and recognition are typically decoupled embedded subsystems.
\end{enumerate}


\subsection{Experimental Protocol}

Five volunteers (three female, two male) participated in our evaluation. For each participant and for each method under comparison, we generated five personalized adversarial patches. Every patch was tested in five independent trials, with the smartphone positioned near the mouth following the setup used in our demonstration. Table~\ref{tab:datasets} summarizes all results.



\begin{table}[h!]
\caption{Comparison with existing adversarial patch approaches in a digital--physical setup. Patches are displayed on a smartphone screen and tested against the Face++ \cite{faceplusplus} system, MobileFaceNet (MF-Net) \cite{chen2018mobilefacenets}, and a real-world FR camera.}
\label{tab:datasets}
\begin{center}
\begin{small}
\begin{tabular}{l|c|c|cc}
\toprule
 & \multicolumn{1}{c}{Face++} & \multicolumn{1}{c}{MF-Net} & \multicolumn{2}{c}{Real Camera} \\
 Method & Sim. $\downarrow$ &  Sim. $\downarrow$ & ASR $\uparrow$ & \makecell{Mean \\ Conf. $\uparrow$} \\
\midrule
 Sibl. Attack \cite{li2023sibling} & 0.804 & 0.20 & 0.24 & 43.9 \\
 FaceOff \cite{FaceOff} & 0.787 & 0.26 & 0.24 & 52.0 \\
 AdvMask \cite{zolfi2023adversarial} & 0.784 & 0.30 & 0.28 & 56.3 \\
 ${DiPA}_{TV}$ (ours) & \textbf{0.755} & \textbf{0.15} & \textbf{0.56} & \underline{61.6} \\
 DiPA (ours)          & \underline{0.777} & \underline{0.17} & \underline{0.48} & \textbf{62.6} \\
\bottomrule
\end{tabular}
\end{small}
\end{center}
\end{table}

\section{Demo}

Figure~\ref{fig:first_figure} illustrates the architecture of the demonstration. The demo highlights a realistic pervasive computing scenario where a user—equipped only with a smartphone—interacts with an intelligent camera system deployed in an environment such as a smart home, office, or access-controlled space.

\textbf{Step 1: User interaction.}
Through a web interface, users upload a portrait photo and provide explicit consent for processing. The back-end service initiates patch generation using the DiPA pipeline.

\textbf{Step 2: Patch creation.}
Depending on system load, the server returns a set of personalized adversarial patches within 10 minutes to 1 hour. These patches are optimized for digital-physical deployment and can be displayed on any standard smartphone.

\begin{figure}[htb]
\begin{center}
\centerline{\includegraphics[width=0.73\columnwidth]{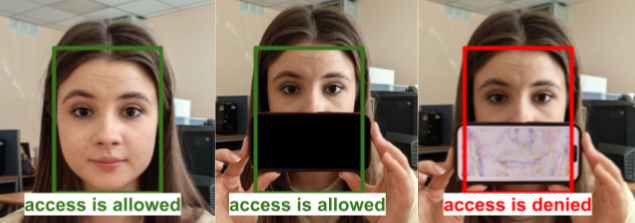}}
\caption{DiPA attack on real-world camera. Note that simply displaying a black or white screen or a patch with a random pattern will still result in the person being correctly detected.}
\label{fig:example}
\end{center}
\end{figure}

\textbf{Step 3: Live physical demonstration.}
Participants hold the smartphone near the lower region of their face, as shown in Figure~\ref{fig:example}.
The real-world FR camera then attempts verification. In successful attacks, the FR system continues to detect the presence of a face but fails to verify the identity of the authorized user—demonstrating a high-impact dodging attack.

The proposed demonstration shows a practical threat in pervasive computing environments where people and devices interact naturally. It illustrates how an everyday smartphone can disrupt deployed face-recognition systems and highlights the need for stronger, context-aware security measures in pervasive, FR-dependent infrastructures.

The demo is fully hands-on, allowing attendees to:
\begin{itemize}
\item observe live FR failures on a commercial camera,
\item visualize differences between digital and digital-physical attack modes,
\item understand the fragility of current sensing systems under dynamic mobile interactions.
\end{itemize}

\section{Conclusion}
This demonstration presents DiPA, a digital-physical adversarial attack that uses a smartphone screen as the attack medium, showing a practical and overlooked threat to face-recognition systems in pervasive computing environments. Through a complete workflow and live tests on a commercial FR camera, we show that DiPA can reliably perform dodging attacks while keeping the face detectable, revealing weaknesses in camera-based authentication systems used in many smart spaces. Our results highlight that the close interaction of mobile devices, sensors, and AI in pervasive computing creates new attack surfaces that standard digital evaluations cannot capture, underscoring the need for stronger, context-aware defenses. Although we do not test proprietary defenses, we have shared our findings with the camera manufacturer. Future work includes exploring adaptive defenses, on-device anomaly detection for mobile–camera interactions, and more resilient FR pipelines designed for pervasive computing ecosystems.

\section*{Acknowledgment}

This work was supported by the The Ministry of Economic Development of the Russian Federation in accordance with the subsidy agreement (agreement identifier 000000C313925P4H0002; grant No 139-15-2025-012).

The research was carried out using the MSU-270 supercomputer of Lomonosov Moscow State University.

\bibliographystyle{IEEEtran}
\bibliography{IEEEabrv, sample-base}

@String{Computing = "Computing" }

@String{Computer = "{IEEE} Computer" }

@String{Springer = "Springer-Verlag" }

@ArtifactSoftware{R,
    title = {R: A Language and Environment for Statistical Computing},
    author = {{R Core Team}},
    organization = {R Foundation for Statistical Computing},
    address = {Vienna, Austria},
    year = {2019},
    url = {https://www.R-project.org/},
}

@inproceedings{li2023sibling,
  title={Sibling-attack: Rethinking transferable adversarial attacks against face recognition},
  author={Li, Zexin and Yin, Bangjie and Yao, Taiping and Guo, Junfeng and Ding, Shouhong and Chen, Simin and Liu, Cong},
  booktitle={Proceedings of the IEEE/CVF conference on computer vision and pattern recognition},
  pages={24626--24637},
  year={2023}
}

@inproceedings{zolfi2023adversarial,
  title={Adversarial Mask: Real-World Universal Adversarial Attack on Face Recognition Models},
  author={Zolfi, Alon and Avidan, Shai and Elovici, Yuval and Shabtai, Asaf},
  booktitle={Machine Learning and Knowledge Discovery in Databases: European Conference, ECML PKDD 2022, Grenoble, France, September 19--23, 2022, Proceedings, Part III},
  pages={304--320},
  year={2023},
  organization={Springer}
}

@misc{FaceOff,
  author = {Ronaldas Paulius Lencevicius},
  howpublished = {GitHub},
  title = {Face-Off: Steps towards physical adversarial attacks on facial recognition},
  URL = {https://github.com/392781/FaceOff},
  month = {Aug},
  year = {2019},
}

@inproceedings{chen2018mobilefacenets,
  title={Mobilefacenets: Efficient cnns for accurate real-time face verification on mobile devices},
  author={Chen, Sheng and Liu, Yang and Gao, Xiang and Han, Zhen},
  booktitle={Chinese conference on biometric recognition},
  pages={428--438},
  year={2018},
  organization={Springer}
}

@inproceedings{mahendran2015understanding,
  title={Understanding deep image representations by inverting them},
  author={Mahendran, Aravindh and Vedaldi, Andrea},
  booktitle={Proceedings of the IEEE conference on computer vision and pattern recognition},
  pages={5188--5196},
  year={2015}
}

@inproceedings{deng2019arcface,
  title={Arcface: Additive angular margin loss for deep face recognition},
  author={Deng, Jiankang and Guo, Jia and Xue, Niannan and Zafeiriou, Stefanos},
  booktitle={Proceedings of the IEEE/CVF conference on computer vision and pattern recognition},
  pages={4690--4699},
  year={2019}
}

@inproceedings{wang2018cosface,
  title={Cosface: Large margin cosine loss for deep face recognition},
  author={Wang, Hao and Wang, Yitong and Zhou, Zheng and Ji, Xing and Gong, Dihong and Zhou, Jingchao and Li, Zhifeng and Liu, Wei},
  booktitle={Proceedings of the IEEE conference on computer vision and pattern recognition},
  pages={5265--5274},
  year={2018}
}

@inproceedings{meng2021magface,
  title={Magface: A universal representation for face recognition and quality assessment},
  author={Meng, Qiang and Zhao, Shichao and Huang, Zhida and Zhou, Feng},
  booktitle={Proceedings of the IEEE/CVF conference on computer vision and pattern recognition},
  pages={14225--14234},
  year={2021}
}

@misc{faceplusplus,
  howpublished = {Online},
  title = {Face++ AI Open Platform},
  URL = {https://www.faceplusplus.com/},
  note = {Accessed: 2025-05-10}
}

@inproceedings{pautov2019adversarial,
  title={On adversarial patches: real-world attack on arcface-100 face recognition system},
  author={Pautov, Mikhail and Melnikov, Grigorii and Kaziakhmedov, Edgar and Kireev, Klim and Petiushko, Aleksandr},
  booktitle={2019 International Multi-Conference on Engineering, Computer and Information Sciences (SIBIRCON)},
  pages={0391--0396},
  year={2019},
  organization={IEEE}
}

@INPROCEEDINGS{8958122,
  author={Kaziakhmedov, Edgar and Kireev, Klim and Melnikov, Grigorii and Pautov, Mikhail and Petiushko, Aleksandr},
  booktitle={2019 International Multi-Conference on Engineering, Computer and Information Sciences (SIBIRCON)}, 
  title={Real-world Attack on MTCNN Face Detection System}, 
  year={2019},
  volume={},
  number={},
  pages={0422-0427},
  doi={10.1109/SIBIRCON48586.2019.8958122}}

\end{document}